\documentclass[10pt,twocolumn,letterpaper]{article}

\usepackage{cvpr}              
\definecolor{cvprblue}{rgb}{0.21,0.49,0.74}
\usepackage[pagebackref,breaklinks,colorlinks,allcolors=cvprblue]{hyperref}

\def\paperID{2499} 
\def\confName{CVPR}
\def\confYear{2026}
\usepackage{times}
\usepackage{epsfig}
\usepackage{graphicx}
\usepackage{amsmath}
\usepackage{amssymb}
\usepackage{multirow}
\usepackage{tikz}
\usepackage{tikz-cd}
\usetikzlibrary{arrows,automata}
\usetikzlibrary{shadows}
\usetikzlibrary{plotmarks}
\usetikzlibrary{shapes}
\usepackage{array}
\usepackage{booktabs} 
\usepackage{colortbl}
\usepackage{xcolor}
\usepackage{soul}
\usepackage{arydshln}
\usepackage{fontawesome5}
\usepackage[normalem]{ulem}

\newcolumntype{?}{!{\vrule width 1pt}}
\newcolumntype{|}{!{\vrule width .5pt}}
\definecolor{darkred}{rgb}{0.6148, 0., 0.}
\definecolor{lightyellow}{rgb}{1., 1., 0.95}
\definecolor{lightgrape}{rgb}{0.93, 0.89, 0.98}

\usepackage{pifont}
\definecolor{cvprblue}{rgb}{0.21,0.49,0.74}
\newcommand*{\audioicon}{\raisebox{-.75mm}{\includegraphics[scale=0.035]{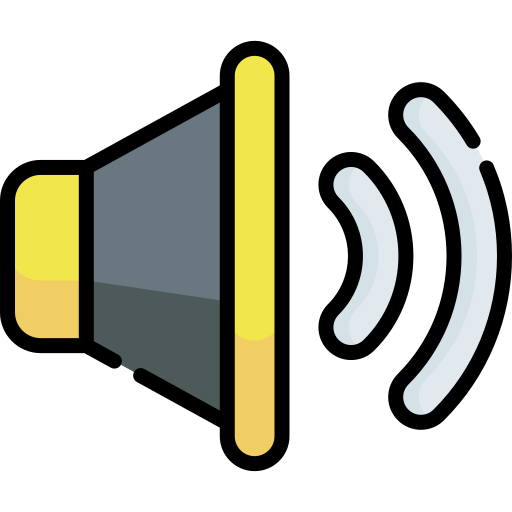}}}
\newcommand*{\visualicon}{\raisebox{-.75mm}{\includegraphics[scale=.035]{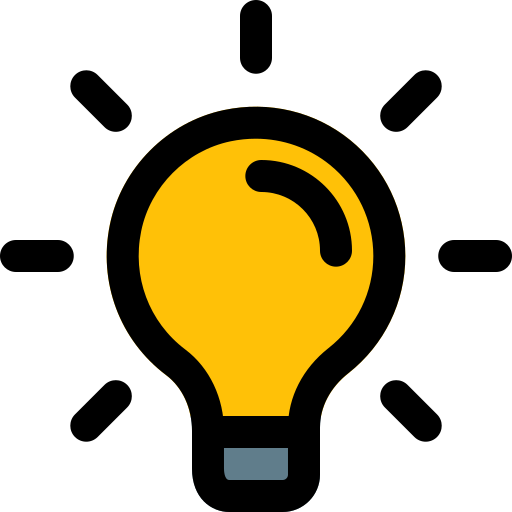}}}
\newcommand*{\texticon}{\raisebox{-1mm}{\includegraphics[scale=.035]{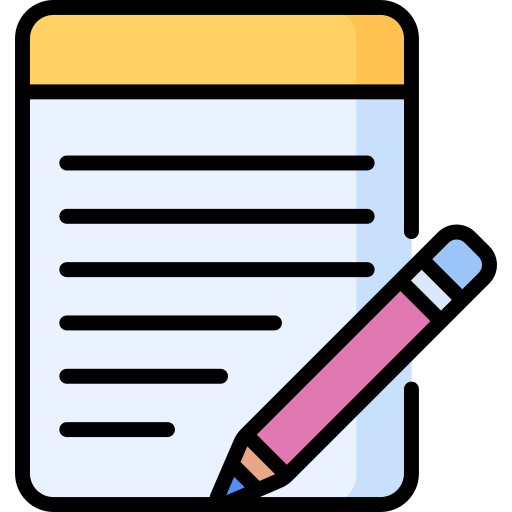}}}
\newcommand*{\pyramidicon}{\raisebox{-1.25mm}{\includegraphics[scale=0.03]{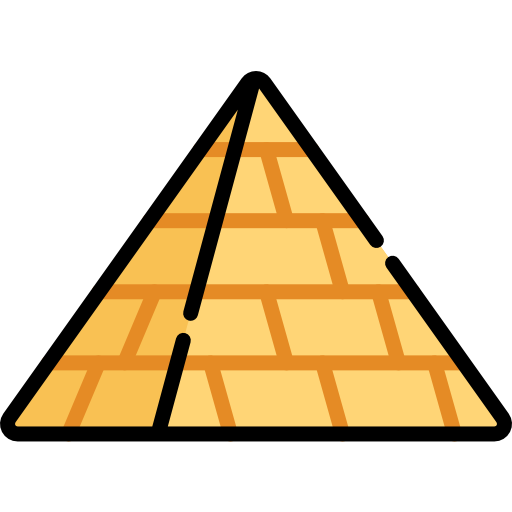}}}

\newcommand*{\audioiconincap}{\raisebox{-.75mm}{\includegraphics[scale=0.028]{images/icons/volume.png}}}
\newcommand*{\visualiconincap}{\raisebox{-.75mm}{\includegraphics[scale=0.028]{images/icons/lamp.png}}}
\newcommand*{\texticonincap}{\raisebox{-1mm}{\includegraphics[scale=0.025]{images/icons/note.png}}}
\newcommand*{\pyramidiconincap}{\raisebox{-1.25mm}{\includegraphics[scale=0.025]{images/icons/pyramid.png}}}

\newcommand{\xdasharrow}[2][->]{
\tikz[baseline=-\the\dimexpr\fontdimen22\textfont2\relax]{
\node[anchor=south,font=\large, inner ysep=1.5pt](x){#2};
\draw[line width = .35mm, dotted,#1](x.south west)--(x.south east);
}
}
\newcommand{\xarrow}[2][->]{
\tikz[baseline=-\the\dimexpr\fontdimen22\textfont2\relax]{
\node[anchor=south,font=\large, inner ysep=1.5pt](x){#2};
\draw[line width = .35mm, #1](x.south west)--(x.south east);
}
}
\title{
\vspace{-10pt}
AuralSAM2: Enabling SAM2 Hear \\ Through Pyramid Audio-Visual Feature Prompting

\vspace{-20pt}
}

\author{
\parbox{\linewidth}{\centering Yuyuan Liu \textsuperscript{\rm 1}  $\quad$ Yuanhong Chen \textsuperscript{\rm 2} $\quad$ Chong Wang \textsuperscript{\rm 3} $\quad$ Junlin Han \textsuperscript{\rm 1} $\quad$ Junde Wu \textsuperscript{\rm 1} $\quad$ \\ Can Peng \textsuperscript{\rm 1} $\quad$ Jingkun Chen \textsuperscript{\rm 1} $\quad$ Yu Tian \textsuperscript{\rm 4}$^{\text{(\faEnvelope[regular])}}$ $\quad$ Gustavo Carneiro \textsuperscript{\rm 5}
\\   \vspace{10pt}
\small  \textsuperscript{\rm 1} Department of Engineering Science, University of Oxford
 $\quad$   \textsuperscript{\rm 2} Australian Institute for Machine Learning, Adelaide University \\
 \textsuperscript{\rm 3} Stanford University  $\quad$  
 \textsuperscript{\rm 4} University of Central Florida  $\quad$  
 \textsuperscript{\rm 5} University of Surrey }}

\begin{document}
\maketitle
\begin{abstract}
Segment Anything Model 2 (SAM2) exhibits strong generalisation for promptable segmentation in video clips; 
however, its integration with the audio modality remains underexplored.
Existing approaches either convert audio into visual prompts (e.g., boxes) via foundation models, or inject adapters into the image encoder for audio–visual fusion. Yet both directions fall short in human-in-the-loop scenarios due to limited prompt accuracy and increased inference overhead.
In particular, these adapter-based methods often suffer from audio prompt dilution, where the signal gradually weakens as it propagates through the network. 
In this work, we propose AuralSAM2, which integrates audio into SAM2 while largely preserving its promptable segmentation capability.
Its core module, AuralFuser, fuses audio and visual features to generate sparse and dense prompts.
Guided by audio and built upon SAM2’s feature pyramid, these prompts propagate auditory cues across visual layers, reinforcing cross-modal influence.
To further align modalities, we introduce an audio-guided contrastive loss that emphasises auditory relevance in dominant visual features.
Our method achieves notable accuracy gains on public benchmarks with only minimal impact on the interactive efficiency of promptable segmentation. 
Our code is available at \url{https://github.com/yyliu01/AuralSAM2}.
\vspace{-10pt}
\end{abstract}

\section{Introduction}
\newcommand{\mystartt}{\hspace{-2.5pt}\raisebox{-0.1 em}{ \tikz{\node[draw=black, color=black, fill=white, star, minimum
width=0.30cm, minimum height=0.30cm,inner sep=1pt, line width=0.35mm, circular drop shadow, star point ratio=2.25,] at (0,0) {};}}\hspace{2.5pt}}
\newcommand{\mycirclet}{\hspace{-2.5pt}\raisebox{-0.1 em}{ \tikz{\node[draw=black, color=black, fill=white, circle, minimum
width=0.25cm, minimum height=0.25cm,inner sep=1pt, line width=0.35mm, circular drop shadow] at (0,0) {};}}\hspace{2pt}}

\newcommand{\YUinline}[1]{%
  {\color{orange}\textbf{YU:}~#1}%
}
Large vision foundation models have emerged as a key advancement in computer vision~\cite{caron2021emerging,caron2021emerging, oquab2023dinov2}, offering versatile and transferable visual representations across domains.
\begin{figure}
    \centering
    \includegraphics[width=\linewidth]{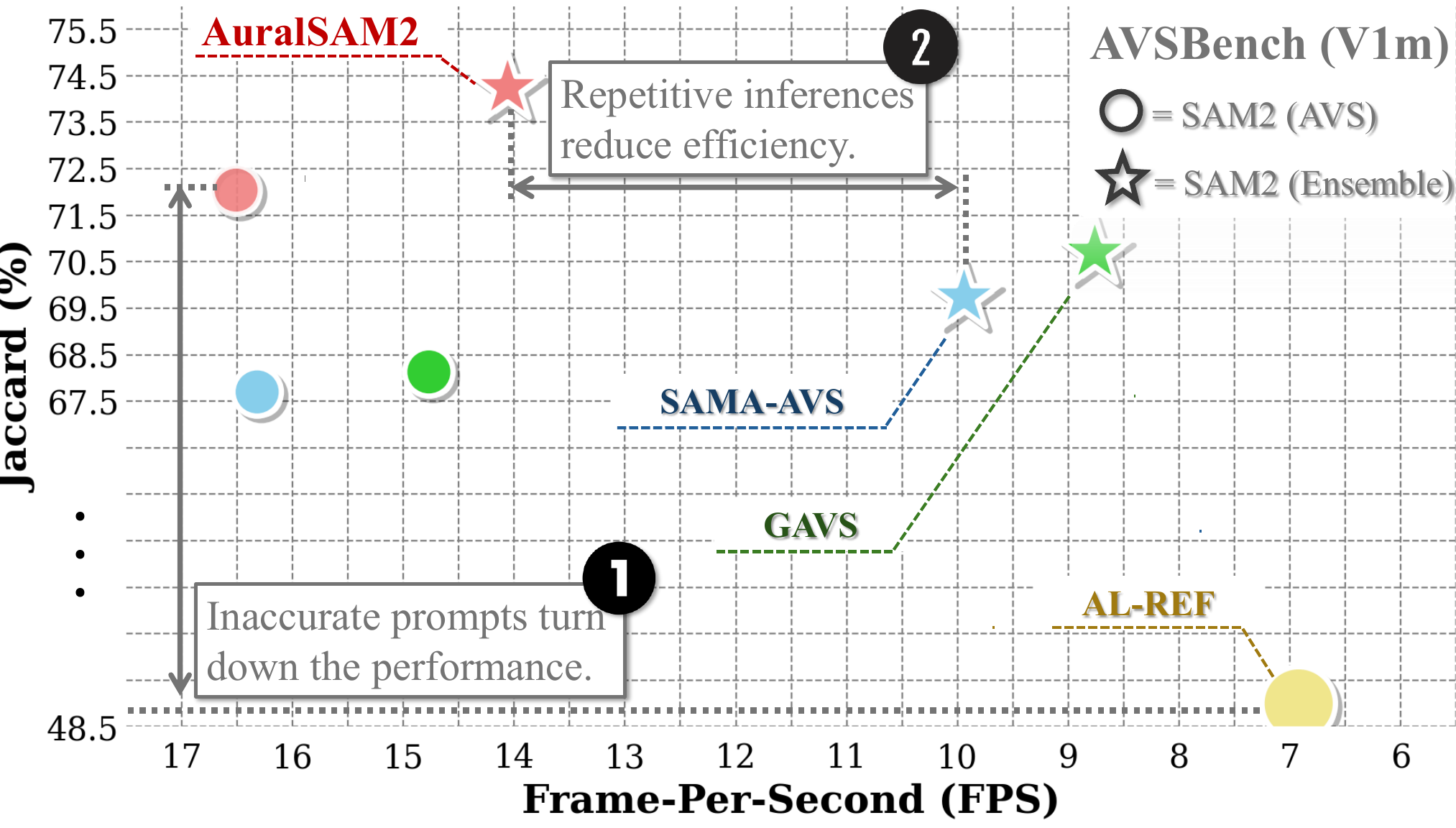}
    \caption{\textbf{Prompt Engineering for Integrating Audio Signal} in AVSBench (V1m)~\cite{zhou2022audio}. \protect\mycirclet SAM2 (AVS) includes adapter-based methods GAVS~\cite{wang2024prompting} and SAMA-AVS~\cite{liu2024annotation}, along with AL-REF~\cite{huang2024unleashing}, which process audio signals to segment sounding objects. 
    To simulate human-in-the-loop scenarios, \protect\mystartt SAM2 (Ensemble) combines the SAM2 (AVS) results with SAM2 outputs guided by point \& box prompts generated from ground truth. } 
    \vspace{-15pt}
    \label{fig: prompt_engineering_first_page}
\end{figure}
Among them, the Segment Anything Model (SAM) series~\cite{kirillov2023segment, ravi2024sam} pioneered promptable segmentation via a human–in-the-loop interactive paradigm.
In particular, SAM2~\cite{ravi2024sam} extends this paradigm to video by propagating human-provided visual prompts (e.g., points, boxes) across frames to segment targets of interest throughout a clip. \\
\indent However, real-world scenarios often require a deeper understanding beyond visual features alone~\cite{xu2023multimodal}. Auditory signals, which frequently coexist with video frames, are not incorporated into SAM2's inherent design~\cite{ravi2024sam}. 
As a result, users are left to manually scrub through video frames to identify sounding targets, such as a speaking person~\cite{shin2024edit,alcazar2022end}, or an anomalous object making noise~\cite{liu2025exploring,koizumi2020description}.
This process is slow~\cite{ding2025sam2long, caelles20182018,heo2020interactive} and error-prone~\cite{wei2024training, vujasinovic2024strike}, especially when the object is small~\cite{hosoya2024rethinking} or visually ambiguous~\cite{schwirten2024ambiguous}.  
In such cases, audio cues serve as a natural guide: they help narrow the search space and stabilise object tracking under occlusion or among look-alike instances.
These advantages highlight the potential of audio guidance in promptable segmentation workflows, leading to the core question:  
\textit{How can we integrate audio guidance into SAM2 without compromising its prompt-driven design for human–AI collaboration?} \\
\begin{figure}
    \centering
    \includegraphics[width=.95\linewidth]{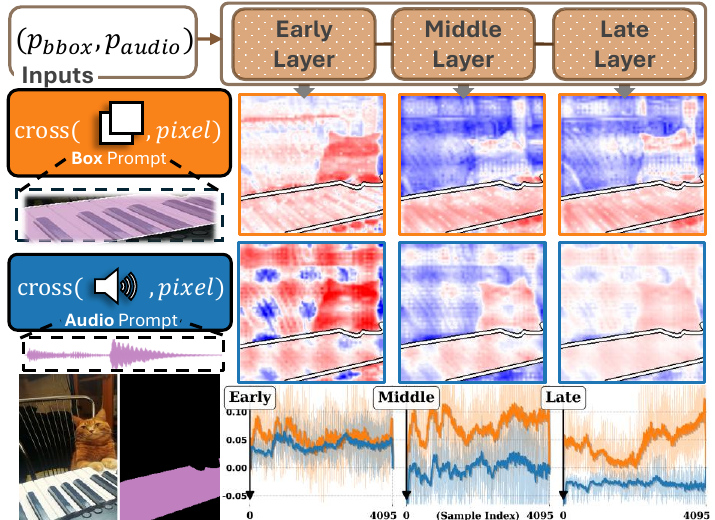}
\caption{
\textbf{Audio Prompt Dilution.}
The audio prompt signal weakens as it propagates through the SAM2 backbone from~\cite{wang2024prompting}. 
The heatmap visualizes audio–visual cross-attention, and the curve traces its pixel-wise intensity.
In contrast, a pretrained bounding box prompt maintains strong alignment throughout the network.}
    \label{fig:placeholder}
    \vspace{-15pt}
\end{figure}
A promising direction is Audio-Visual Segmentation (AVS)~\cite{zhou2022audio}, which explores the semantic relationships between audio and pixel-level visual features in video clips.  
One common approach~\cite{huang2024unleashing, chen2025openavs, zhou2025think} is to leverage multimodal foundation models to translate audio into textual descriptions, which are then used to generate visual prompts for SAM2 to localise sounding objects.  
However, as illustrated in Fig.~\ref{fig: prompt_engineering_first_page} (\ding{182}), taken from AL-REF~\cite{huang2024unleashing},
such generated prompts often suffer inaccuracies from hallucination~\cite{liu2024survey}. For instance, a box prompt may produce a mask that captures internal patterns instead of the object itself.  
Moreover, reliance on foundation models increases inference latency and incurs additional costs due to API-based querying~\cite{zhou2025think}.
\\
\indent 
Another line of research~\cite{seon2024extending, jia2022visual} introduces audio guidance to SAM2 by injecting adapters into its image encoder, enabling audio–visual feature fusion.
However, this integration \emph{alter the intermediate visual features} and degrades SAM2’s promptable segmentation performance.
In prompt engineering scenarios with human-in-the-loop, as illustrated in Fig.~\ref{fig: prompt_engineering_first_page} (\ding{183}), these methods~\cite{wang2024prompting, liu2024annotation} require \textbf{repeated SAM2 inferences}:  
one forward pass to process and fuse audio signals via the adapters (producing audio-conditioned visual features), and another to handle human-provided prompts through SAM2’s promptable interface.  
This repeated inference significantly slows down the system.
For example, \mystartt ensemble results from~\cite{wang2024prompting, liu2024annotation} are nearly 6.5 FPS slower than their \mycirclet AVS results, affecting its real-time feedback performance in practice. 
\\
\indent 
More critically, 
unlike task-specific methods~\cite{chen2025ccstereo, gao2023avsegformer} that tightly couple audio and vision via end-to-end training, adapter-based methods retain a frozen (SAM) backbone and rely on minimal trainable components. 
This shift poses a unique challenge: audio is not inherently compatible with SAM’s prompt-based design. 
Comparing with visual prompts, it lacks spatial anchoring and unfolds on a different temporal scale. Simply injecting adapters~\cite{wang2024prompting, liu2024annotation} offers limited control over how audio and pixel signals are fused and propagated across layers. Worse still, the decoder is overwhelmingly dominated by visual features: a single clip yields over \(10^6\) dense visual tokens, while audio contributes only around 10 coarse embeddings. Taken together, these factors lead to a phenomenon we term \ul{\textit{audio prompt dilution}}: as attention propagates deeper into the model, audio guidance progressively fades. As shown in Fig.~\ref{fig:placeholder}, while the box prompt maintains strong cross-attention signal with pixel features throughout the decoder, the post-trained audio prompt from~\cite{wang2024prompting} weakens progressively, losing its cross-modal correspondence. This is not merely under-utilised audio; it reflects a structural mismatch between how prompts are expected to function in SAM and what audio, in its current form, can reliably deliver in human-in-the-loop workflows.
\\
\indent In this work, we propose \textit{AuralSAM2}, a method designed to enrich SAM2 with audio guidance without compromising its prompt-driven interface. At the core of our method is the AuralFuser module, which is externally attached to the frozen SAM2. This design allows the model to perceive audio signals without modifying image features, thereby avoiding repeated inferences in prompt engineering. 
To mitigate audio prompt dilution, AuralFuser enhances audio-conditioned attention by generating two complementary sets of feature-level prompts: 
sparse prompts that capture high-level contextual cues  of potential sounding objects, while dense prompts ensure precise pixel-level alignment. These prompts are progressively derived by aligning audio features with a multi-scale feature pyramid built upon patch embeddings from SAM2.
This hierarchical design preserves audio guidance throughout the network and strengthens its influence on segmentation.
To further counter visual dominance, we introduce an audio-guided contrastive learning (AudioCon) strategy. AudioCon pulls relevant visual features  (from pyramid) toward audio prototypes while ignoring visual–visual pairs, reinforcing auditory influence in cross-modal alignment.
To summarise, our AuralSAM2's contributions are:
\begin{itemize}
    \item We propose AuralFuser, a module that generates audio-conditioned prompts without modifying SAM2’s visual backbone, enabling efficient promptable inference;
    \item To mitigate audio prompt dilution, AuralFuser constructs sparse and dense prompts through feature pyramid integration, ensuring auditory signal is preserved; and
    \item We propose AudioCon to further enhance the alignment between audio signals with hierarchical visual features while mitigating the issue of visual dominance.
\end{itemize}
Our method enables SAM2 to process audio (and optionally language-based audio cues) with minimal efficiency overhead in prompt engineering scenarios. 
As shown in Fig.~\ref{fig: prompt_engineering_first_page}, AuralSAM2 incurs only a 2.3 FPS drop when adapting visual prompts for the mask decoder, while achieving an Jaccard improvement of 3.9\% on AVSBench (V1m)~\cite{zhou2022audio}, outperforming other SAM2-based SOTA methods.


\section{Related Work}
\noindent \textbf{Vision Foundation Model} methods utilise millions of images and rely on self-supervised learning~\cite{caron2021emerging, oquab2023dinov2, simeoni2025dinov3} to enhance feature representation. A notable departure from this trend is the SAM series~\cite{kirillov2023segment, ravi2024sam}, which introduces a semi-automated, human-in-the-loop training paradigm. By expanding labeled data through self-generated or human-refined visual prompts (e.g., points and boxes), SAM learns diverse visual patterns across both static images~\cite{kirillov2023segment} and video clips~\cite{ravi2024sam}. 
In this work, our method is built upon SAM2, chosen for its video-specific design and its strong promptable segmentation capabilities, which we aim to extend to the audio modality without sacrificing human-in-the-loop efficiency.
\\
\noindent \textbf{Audio–Visual Learning (AVL)} has been widely studied in deep learning to uncover semantic relationships between audio and visual modalities for enhanced machine perception~\cite{zhu2021deep}. It includes tasks such as source separation~\cite{liu2024separate, chen2022zero}, which extracts distinct sounds from a mixture; binaural audio generation~\cite{chen2025ccstereo}, which creates spatial sound from mono or stereo inputs; and sound source localisation~\cite{chen2021localizing, mo2022localizing}, which estimates the direction and distance of sound sources. Despite these advances, modeling pixel-level interactions between the two modalities remains a major challenge.\\
\noindent \textbf{Audio–Visual Segmentation (AVS)} has recently been developed to tackle this challenge, with AVSBench~\cite{zhou2022audio, zhou2023audio} serving as the first benchmark, covering both single and multiple sounding sources. The task has since expanded to include zero-shot segmentation for unseen and unheard objects~\cite{wang2024prompting}, as well as language-aided AVS incorporating textual guidance~\cite{wang2025ref}. \ul{\textit{Task-specific AVS models}} remain the mainstream approach, with networks retrained from scratch on the AVSBench dataset~\cite{zhou2022audio, zhou2023audio}. Most methods focus on cross-modal fusion, aligning visual features with audio signals before feeding them into a transformer decoder~\cite{huang2025revisiting, li2024selm, huang2023discovering, mao2023multimodal}, either directly~\cite{mao2023multimodal, lin2023vision} or through learnable audio queries~\cite{huang2023discovering, li2023catr}. To further improve alignment, \cite{hao2023improving} reconstructs audio embeddings from associated visual features, while~\cite{li2023catr} incorporates temporal cues to enhance spatial correlations between modalities. Contrastive learning~\cite{chen2024unraveling, chen2025cpm} has also been explored to strengthen audio-visual associations in the latent space.
However, these task-specific AVS models~\cite{li2024selm, lin2023vision, ma2024steppingstones} are typically trained on narrow domains, which restricts their generalisability. 
\noindent \ul{\textit{AVS for the SAM series}} is a promising yet underexplored direction that builds on SAM’s strong generalisation. Existing methods mainly integrate audio via adapters~\cite{jia2022visual}, either in the image encoder~\cite{mo2023av, seon2024extending} or across the full architecture~\cite{wang2024prompting}, enabling fine-tuning on AVS datasets. SAMA-AVS~\cite{liu2024annotation} retrains the mask decoder with audio adapters, while GAVS~\cite{wang2024prompting} and AV-SAM~\cite{mo2023av} use audio-visual features as decoder prompts. These approaches modify image features during audio integration, introducing extra inference steps that reduce efficiency. Alternatively, AL-Ref~\cite{huang2024unleashing} and SAM4AVS~\cite{yu2023can} use large language or vision-language models~\cite{achiam2023gpt, liu2024grounding} to extract audio semantics and generate visual prompts in a zero-shot manner, though they often suffer from limited accuracy and slow inference.
Motivated by these limitations, our proposed AuralFuser integrates audio as an external module without altering the features in the image encoder, thereby avoiding the need for repetitive inference. In addition, our method eliminates reliance on external foundation models by directly generating two sets of feature-level prompts through cross-modal fusion. These prompts effectively guide the SAM2 decoder in capturing sounding objects with both high precision and computational efficiency. Building on this design, AudioCon further enhances audio–visual alignment by reducing visual dominance impact and reinforcing the guiding role of audio cues via contrastive learning.

\section{Method}
\begin{figure*}
    \centering
    \includegraphics[width=.97\linewidth]{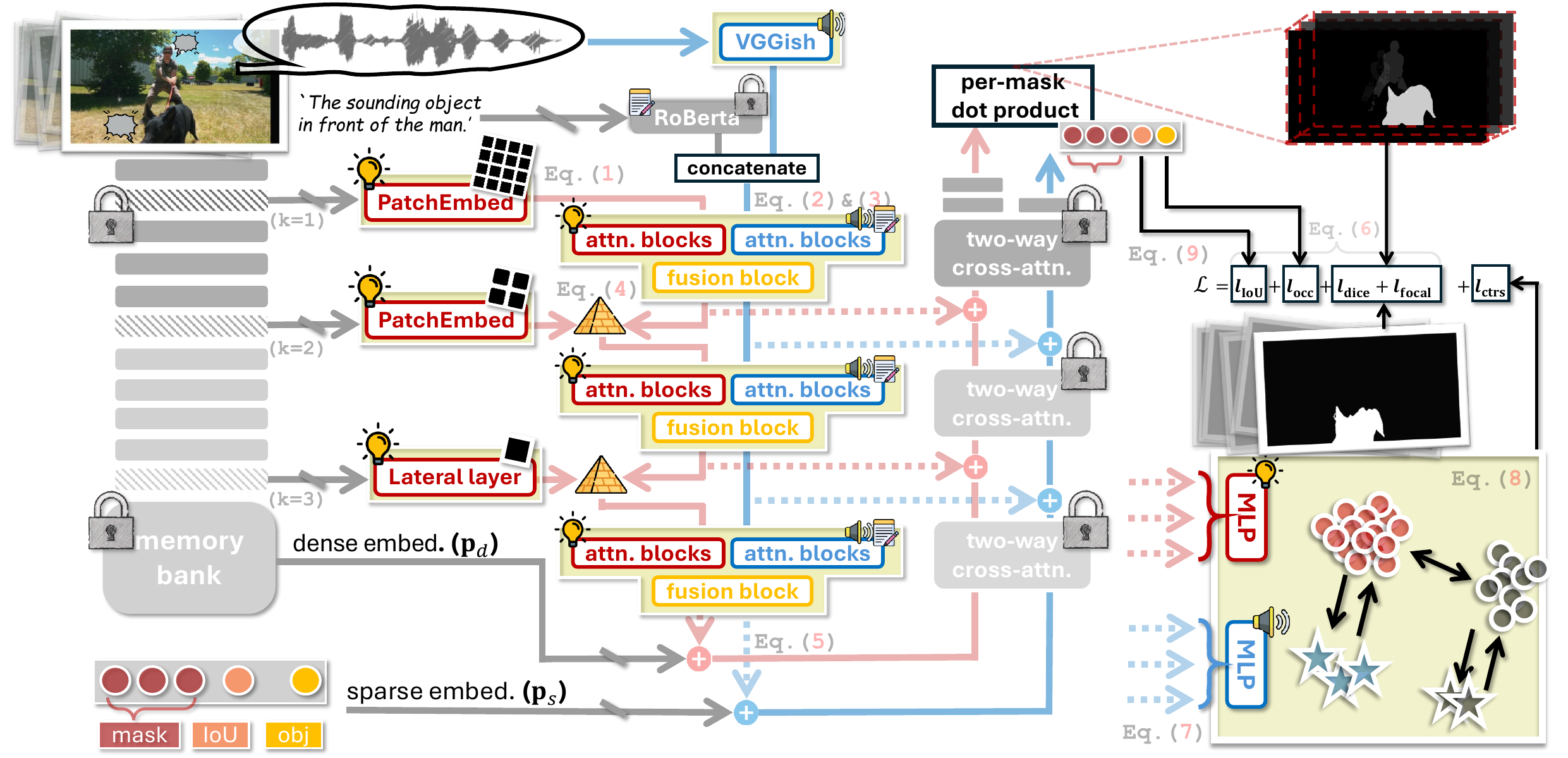}
    \vspace{-5pt}
    \caption{\textbf{Illustration of our approach in a language-aided AVS dataset~\cite{wang2025ref}}. Audio WAV and text sentences are processed via VGGish~\cite{chen2020vggsound} and RoBERTa~\cite{liu2019roberta}, respectively, and then combined. Visual features are extracted from SAM2 in a pyramid structure and processed through PatchEmbedding in Eq.~\eqref{eq:patch_embed} with varying patch sizes (equivalent to the Lateral Layer when \(\texttt{k=3}\)), then merged using Eq.~\eqref{eq:smooth}. The visual and audio-text features then undergo self-attention from Eq.~\eqref{eq:self_attn} and fusion blocks in Eq.~\eqref{eq:fusion} to generate sparse and dense feature-level prompts, which guide the mask decoder in capturing potential sounding objects, constrained by the SAM2 loss in Eq.~\eqref{eq:sam2_loss} and audio-guided CL (AudioCon) in Eq.~\eqref{eq:ctrs_loss}. Please note that operations based on fused features are highlighted using \textcolor{darkred!50}{\textbf{\xdasharrow{$\;$}}} and \textcolor{blue!30}{\textbf{\xdasharrow{$\;$}}}.
    }
    \vspace{-15pt}
    \label{fig:workflow}
\end{figure*}
We define the language-aided AVS dataset~\cite{wang2025ref} as $ 
    \mathcal{D} = \left\{ \left( \mathbf{a}_i, \mathbf{t}_i, \mathbf{v}_i \right) \mid \mathbf{v}_i = \left\{ (\mathbf{x}_{ij}, \mathbf{y}_{ij}) \right\}_{j=1}^{B} \right\}_{i=1}^{|\mathcal{D}|},$
where \(|\mathcal{D}|\) denotes the number of video clips.  The audio signal $\mathbf{a}_i \in \mathcal{A} \subset \mathbb{R}^{N^a \times 2}$ represents a waveform,
with \(N^a\) being the duration of the audio (based on $16000$ Hz sampling rate) with 2 channels.   
The expression text \(\mathbf{t}_i \in \mathcal{T} \subset \mathbb{R}^{1\times N^t} \) denotes a sentence with $N^t$ words.
Each video sequence \(\mathbf{v}_i\) consists of \(B\) pairs of RGB image $\mathbf{x}_{ij} \in \mathcal{X} \subset \mathbb{R}^{H \times W \times 3},$ 
with a spatial resolution of \(H \times W\), and corresponding pixel-level binarized ground truth masks $\mathbf{y}_{ij} \in \mathcal{Y} \subset [0, 1]^{H \times W}$, representing the sounding object in frame  $j \in\{1,...,B\}$. Note that in some AVS datasets~\cite{zhou2022audio, zhou2023audio}, the language modality \(\mathcal{T}\) is unavailable, in which case our work relies solely on audio and visual modalities.

\subsection{Preliminaries: SAM2}
\label{sec:3_1_sam2_pre}
We define the whole SAM2 as \(\mathbf{f}^{\phi}_{\text{SAM2}}: \mathcal{X} \xrightarrow{\{\mathbf{p}_s, \mathbf{p}_d\}} \mathcal{Y}\), parameterised by \(\phi\), where \(\mathbf{p}_s \in \mathbb{R}^{B \times 5 \times L}\)  represents 5 output tokens of dimension \(L\) and 
\(\mathbf{p}_d \in \mathbb{R}^{B \times H' \times W' \times L}\) denotes the dense feature maps.
Specifically, \(\mathbf{p}_s\) comprises 3 mask tokens, 1 object token, and 1 Intersection-Over-Union (IoU) token. Typically, these tokens are concatenated with sparse prompt embeddings (e.g., from points and boxes). 
The dense features \(\mathbf{p}_d\) are computed as the sum of dense (mask) prompt embeddings and visual features, with an output  resolution \(H'=\frac{H}{16}\) with \(W'=\frac{W}{16}\). 
Since we do not utilise any of the SAM's prompts in the training, we simplify notation by referring to $\mathbf{p}_s$ as the sparse embeddings and $\mathbf{p}_d$ as the dense embedding in the following discussion.\\
SAM2 is composed of an image encoder represented by \(\mathbf{h}^{\phi_h}_{\text{SAM2}}: \mathcal{X} \xrightarrow{} \mathcal{Z}_v\), a memory bank that regularizes the latent feature \(\mathcal{Z}_v\), and a mask decoder \(\mathbf{g}^{\phi_g}_{\text{SAM2}}: \mathcal{Z}_v \xrightarrow{\{\mathbf{p}_s, \mathbf{p}_d\}} \mathcal{Y}\), such that \(\mathbf{f}^{\phi}_{\text{SAM2}} = \mathbf{h}^{\phi_h}_{\text{SAM2}}  \circ \mathbf{g}^{\phi_g}_{\text{SAM2}} \).
In the mask decoder \(\mathbf{g}^{\phi_g}_{\text{SAM2}}\), two-way cross-attention blocks between \(\mathbf{p}_s\) and \(\mathbf{p}_d\) occur 3 times, with the sparse and dense features at each block defined as \(\mathbf{G} = \left\{ \mathbf{p}_{sk}, \mathbf{p}_{dk} | k \in \{ 1, 2, 3 \} \right\}\). After processing the final set (\(k=3\)) of these tokens through three successive MLPs, the group of predicted binarised masks is computed with the following dot product per mask: \(\hat{y}^{\text{mask}} =  \mathbf{p}_{d3} \cdot \mathbf{p}_{s3}^{\text{mask}} \in \mathcal{Y}\).  The predicted  \(\hat{y}^{\text{obj}} \in \mathbb{R}\) is a logit derived from \(\mathbf{p}_{s3}^{\text{obj}}\) to classify the presence of the target in the current scene. The IoUs of the predicted masks, denoted by \(\hat{y}^{\text{IoU}} \in [0,1]\) are obtained from \(\mathbf{p}_{s3}^{\text{IoU}}\) to estimate the overall quality of the output \(\hat{y}^{\text{mask}}\). 

\subsection{AuralFuser}
As shown in Fig.~\ref{fig:workflow}, AuralFuser processes multi-modal features using pre-trained models as follows:
\begin{itemize}
\itemsep0em 
    \item[\audioicon]  The \textit{audio waveform} is compressed via \(\mathbf{f}^{\theta^\texttt{vgg}}_{\texttt{\textbf{VGG}}}: \mathcal{A} \xrightarrow{} \mathcal{Z}_a\),
    where \(\mathbf{z}_a \in \mathcal{Z}_a \subset  \mathbb{R}^{B\times L}\) and \(\theta^{\texttt{vgg}}\) denotes the parameter of  VGGish~\cite{chen2020vggsound}; 
    \item[\texticon] The \textit{textual expression} is processed via \(\mathbf{f}^{\psi}_{\texttt{\textbf{Roberta}}}: \mathcal{T} \xrightarrow{} \mathcal{Z}_t\), where \(\mathbf{z}_t \in \mathcal{Z}_t \subset  \mathbb{R}^{N^t\times L}\) and \(\psi\) denotes the parameter of RoBerta~\cite{liu2019roberta}; and
    \item[\visualicon] The \textit{visual features} are extracted after Q-pooling layers~\cite{ryali2023hiera} to build the pyramid, defined as
    \({\mathbf{Z}_v = \{\mathbf{z}_v^{\texttt{(k)}} \in \mathbb{R}^{B \times \frac{H}{\textbf{s}^{\texttt{(k)}}} \times \frac{W}{\textbf{s}^{\texttt{(k)}}} \times L} \mid \textbf{s}^{\texttt{(k)}} \in \{4, 8, 16\}, \;}\) \({k \in \{1,2,3\} \}}\), with \(\mathbf{Z}_v \subset \mathcal{Z}_v\).  
\end{itemize}
During training, we only update parameters \(\theta\) (e.g., \(\theta^\texttt{vgg}\) as in~\cite{chen2025cpm, chen2024unraveling}), while keeping the text model parameters 
\(\psi\)
and SAM2 parameters \(\phi = \{\phi^g, \phi^h\}\) fixed.
Next, we concatenate the audio and text features to form \(
\mathbf{z}_c = \left[ \mathbf{z}_a, \mathbf{z}_t \right]\), where \(\mathbf{z}_c \in \mathbb{R}^{(B+N^t)\times L}\) and apply subsequent operations within our framework that are explained below.

\noindent \textbf{Pyramid Processing:} for each $k \in \{1,2,3\}$, we process the visual features as follows:
\begin{equation}
    \mathbf{\tilde{z}}_v^{\texttt{(k)}} = \textbf{f}_\texttt{\textbf{PatchEmbed}}^{\texttt{(k)}}(\mathbf{z}_v^{\texttt{(k)}}; \theta_{pe}^{\texttt{(k)}}, p^{\texttt{(k)}}), \quad p^{\texttt{(k)}} \in \{4,2,1\},
    \label{eq:patch_embed}
\end{equation}
where 
\(\textbf{f}_\texttt{\textbf{PatchEmbed}}^{\texttt{(k)}} (\ \cdot\ ; \theta_{pe}^{\texttt{(k)}}, p^{\texttt{(k)}})\) denotes the patch embedding layer with patch size $(p^{\texttt{(k)}} \times p^{\texttt{(k)}})$  to project all  features to the same resolution with $\mathbf{z}_v^{\texttt{(k)}} \in \mathbb{R}^{B\times H' \times W' \times L}$, and it is  equivalent to the Lateral Layer when \texttt{\small k=3} in previous FPN study~\cite{lin2017feature}.
Self-attention is then applied independently to both modalities:
\begin{equation}
    \begin{aligned}
        \mathbf{r}_c^{\texttt{(k)}} &=  \textbf{f}_{\texttt{\textbf{Attn}}^\textbf{c}}^{\texttt{(k)}}(\mathbf{z}_c + \texttt{\small Pos}^c;\theta^{\texttt{(k)}}_c), \\
        \mathbf{r}_v^{\texttt{(k)}} &=  \textbf{f}_{\texttt{\textbf{Attn}}^\textbf{v}}^{\texttt{(k)}}(\mathbf{z}_v^{\texttt{(k)}} + \texttt{\small Pos}^v;\theta^{\texttt{(k)}}_v), 
        \label{eq:self_attn}
\end{aligned}
\end{equation}
where \(\textbf{f}_{\texttt{\textbf{Attn}}^\textbf{c}}^{\texttt{(k)}}(\ \cdot\ ; \theta^{\texttt{(k)}}_a)\) and \(\textbf{f}_{\texttt{\textbf{Attn}}^\textbf{v}}^{\texttt{(k)}}(\ \cdot\ ; \theta^{\texttt{(k)}}_v)\) are the self-attention blocks for the combined audio-text and visual modalities, respectively, with \(\texttt{\small Pos}^a \in \mathbb{R}^{(B+N^t)\times L}\) and \(\texttt{\small Pos}^v\in \mathbb{R}^{B\times H' \times W' \times L}\) denoting their position encodings.
Finally, we perform cross-modal fusion as shown below:
\begin{equation}
    \begin{aligned}
        \textbf{r}_c^{\texttt{(k)}}, \textbf{r}_v^{\texttt{(k)}} &= \textbf{f}_{\texttt{\textbf{CrossFusion}}}^{\texttt{(k)}}(\textbf{r}_c^{\texttt{(k)}} + \texttt{\small Pos}^c, \textbf{r}_v^{\texttt{(k)}} + \texttt{\small Pos}^v;\theta^{\texttt{(k)}}_f), 
        \label{eq:fusion}
    \end{aligned}
\end{equation}
where \(\textbf{f}_{\texttt{\textbf{CrossFusion}}}^{\texttt{(k)}}(\ \cdot\ ; \theta^k_f)\) represents the cross-modality fusion block, adapted from TPAVI~\cite{zhou2022audio} and the two-way cross-attention fusion mechanism (please see more details in the Supp. Section \textcolor{red}{1.3}). \\
For $\texttt{k}\geq2$, we construct the feature pyramid to integrate early fusion results with late-stage cross-modal alignment, demonstrated as `\smash{\pyramidicon}' \hspace{-2pt}  in Fig.~\ref{fig:workflow}, using:
\begin{equation}
    \mathbf{\tilde{z}}_v^{\texttt{(k)}} = \textbf{f}_\texttt{\textbf{Smooth}}^{\texttt{(k)}}(\mathbf{r}_v^{\texttt{(k-1)}} + \mathbf{\tilde{z}}_v^{\texttt{(k)}}; \theta^{\texttt{(k)}}_s),
    \label{eq:smooth}
\end{equation}
where $\textbf{f}_\texttt{\textbf{Smooth}}^{\texttt{(k)}}(\ \cdot\ ; \theta^k_s)$ denotes the convolutional smoothing layer with kernel size equal to 1 and is commonly used in the feature pyramid related works~\cite{lin2017feature, zhao2017pyramid}.
As a result, our approach provides two sets of feature-level prompts.
1) \ul{Sparse prompts} represent visual-language informed audio features 
\(\mathbf{R}_a = \left\{ \mathbf{r}_a^{\texttt{(k)}} = \operatorname{Select}_a\bigl(\mathbf{r}_c^{\texttt{(k)}}\bigr) \in \mathbb{R}^{B \times L} \mid k \in \{1, 2, 3\} \right\}
\), where \(\operatorname{Select}_a(\cdot)\) is the function that extracts the audio feature \(\mathbf{r}_a^\texttt{(k)}\) from the combined representation \(\mathbf{r}_c^{\texttt{(k)}}\), based on its original position from \(\mathbf{z}_a\) in \(\mathbf{z}_c\). 
These features encode global context by capturing the visual data relevant to audio and language modalities.
2) \ul{Dense prompts} correspond to audio-language enriched visual features
\(\mathbf{R}_v = \left\{\textbf{r}_v^{\texttt{(k)}} \in \mathbb{R}^{B\times H' \times W' \times L} \mid k \in \{ 1, 2, 3 \} \right\}\), which provides pixel-level identification of all potential sounding objects within the scene.
\\
\textbf{Hierarchical Prompting.} We progressively integrate the prompt sets \(\mathbf{r}_a^{\texttt{(k)}}\) and \(\textbf{r}_v^{\texttt{(k)}}\) during the two-way cross-attention blocks in \(\mathbf{g}^{\phi_g}_{\text{SAM2}}\) as follows:
\begin{equation}
    \begin{aligned}
    & \mathbf{\tilde{p}}_{sk}^{mask} = \mathbf{p}_{sk}^{mask} + \textbf{r}_a^{\texttt{(k)}},  \quad \mathbf{\tilde{p}}_{sk} \in \mathbb{R}^{B \times 5 \times L}, \\ 
    & \mathbf{\tilde{p}}_{dk} = \mathbf{p}_{dk} + \textbf{r}_v^{\texttt{(k)}}, \quad \mathbf{\tilde{p}}_{dk} \in \mathbb{R}^{B\times H' \times W' \times L}, 
    \end{aligned}
    \label{eq:prompting}
\end{equation}
where \(\mathbf{G} = \left\{ (\mathbf{\tilde{p}}_{sk}, \mathbf{\tilde{p}}_{dk}) \mid \; k \in \{ 1,2,3 \} \right\}\) and we only update the mask token $\mathbf{p}_{sk}^{mask}$ and $\mathbf{p}_{dk}$ in $\mathbf{g}^{\phi_g}_{\text{SAM2}}$. While the other tokens (i.e., $\mathbf{p}_{s}^{\text{IoU}}$, $\mathbf{p}_{s}^{\text{object}}$) can still learn to capture the correct feature via self-attention blocks in \(\mathbf{h}_\text{SAM2}^{\phi_h}\). 
As a result, we follow the training pipeline in SAM2 with the loss: 
\begin{equation}
\begin{aligned}
 \ell_{\text{SAM2}}&(\mathcal{D}, \theta^{\texttt{vgg}}, \theta^\texttt{(k)})  = \ell_{\text{focal}}(\hat{y}^\text{mask}, \mathbf{y}) + \ell_{\text{dice}}(\hat{y}^\text{mask}, \mathbf{y}) \\
& + \ell_{\text{IoU}}\left(\hat{y}^{\text{IoU}}, \texttt{\small \textbf{IoU}}(\hat{y}^\text{mask}, \mathbf{y})\right) 
+ \ell_{\text{occ}}\left(\hat{\mathbf{y}}_{\text{obj}}, \mathbb{I}(\mathbf{y} > 0)\right),
\label{eq:sam2_loss}
\end{aligned}
\end{equation}
where 
\(\hat{y}^\text{mask}\), \(\hat{y}^\text{obj}\) and \(\hat{y}^\text{IoU}\) are defined in the Preliminaries section
,
\(\mathbb{I}(\mathbf{y} > 0) \in \{0, 1\}\) is a binary indicator determining the presence of a foreground object in the label \(\mathbf{y}\), 
and \texttt{\small \textbf{IoU}} represents the IoU calculation metric. For further details on this loss, we refer to the SAM2 paper~\cite{ravi2024sam}.

\subsection{Audio-guided CL (AudioCon)}
Unlike previous contrastive objectives that treat both modalities symmetrically, AudioCon \emph{privileges} audio as the anchor and only repels visual negatives. This design directly addresses the visual dominance observed in SAM2, ensuring that the most salient clusters in the latent space are organised around audio cues rather than purely visual similarities.
In particular, we utilise two MLPs to project the entire feature sets of $\mathbf{R}_a$ and $\mathbf{R}_v$ into the same embedding space with:
\begin{equation}
    \begin{aligned}
        \textbf{e}_a  = \textbf{f}_{\texttt{\textbf{proj}}^a}(\textbf{r}_a^{\texttt{(k)}}; \theta_{pa}),   \quad 
        \textbf{e}_v  = \textbf{f}_{\texttt{\textbf{proj}}^v}(\textbf{r}_v^{\texttt{(k)}}; \theta_{pv}),  
    \end{aligned}
\end{equation}
where the audio modality embedding $\textbf{e}_a \in \mathbb{R}^{B \times C}$ contains frame numbers ($B$) of embedding features, each with dimension $C$. The visual modality embedding $\textbf{e}_v \in \mathbb{R}^{B \times H' \times W' \times C}$
has a significantly larger number of embedding features compared to the audio modality, with $B \times H' \times W' \gg B$.
Based on the label $\textbf{y}$, we thus can construct the audio embedding set $\mathcal{E}_a = \left \{ (\textbf{e}^a_b, \textbf{y}_b) \mid  \  b = 1, 2, ... B  \right \}$; and
similarly, we can construct the visual embedding set $\mathcal{E}_v = \left \{ (\textbf{e}^v_b, \textbf{y}^{(\mathcal{\omega})}_b) \mid \  b = 1, 2, ... B) \right \}$, where 
$\Omega$ is the lattice of ground truth and \(\omega\) denotes a pixel-level position with  
$\omega \in \Omega \subset \mathbb{R}^{H' \times W'}$. Thus, the AudioCon is defined as:
\begin{equation}
\resizebox{\hsize}{!}{$
\begin{aligned}
\ell_{\text{ctrs}}(\mathcal{D}, & \theta_{pa}, \theta_{pv})
= \frac{1}{|\mathcal{E}_v|}\frac{1}{B}\sum_{(\mathbf{e},\mathbf{y}_b^{(\omega)}) \in \mathcal{E}_v} \sum_{\substack{(\mathbf{e}^+,\mathbf{y}_b) \in \mathcal{E}_a \\ \mathbb{I}(\mathbf{y}_b = \mathbf{y}_b^{(\omega)})}} \\
& -\log \frac{\exp\left(\mathbf{e}\cdot\mathbf{e}^+/\tau\right)}
{\exp\left(\mathbf{e}\cdot\mathbf{e}^+/\tau\right) + \sum_{\substack{(\mathbf{e}^-,\mathbf{y}_b^{(\omega)^-}) \in \mathcal{E}_v \\ \mathbb{I}(\mathbf{y}_b^{(\omega)^-} \neq \mathbf{y}_b^{(\omega)})}} \exp\left(\mathbf{e}\cdot\mathbf{e}^-/\tau\right)}.
\end{aligned}
$}
\label{eq:ctrs_loss}
\end{equation}
where $\tau$ is a temperature parameter and $\mathbb{I}(\cdot)$ indicates whether there is a (pixel-level) foreground object matching the current frame's audio. Unlike previous works~\cite{chen2024unraveling, chen2025cpm} that apply InfoNCE~\cite{oord2018representation} to the entire latent space (i.e., $\mathcal{E}_v \bigcup \mathcal{E}_a$), our AudioCon mitigates modality imbalance by pulling visual embeddings toward relevant audio \(\mathbf{e}^+\) while pushing them away from other visual samples \(\mathbf{y}_b^{(\omega)^-} \). This implementation prevents the model from overemphasizing attraction between pixel-level visual embeddings in \(\mathcal{E}_v\). Instead, it aggregates visual features using audio embeddings as central prototypes, thereby ensuring that visual features cluster around meaningful auditory cues. We include t‑SNE visualisations in Supp. Section \textcolor{red}{4.1} to show this effect.

\subsection{Training Objective}
The training of our AuralSAM2 minimises the following loss function:
\begin{equation}
    \begin{aligned}
        \mathcal{L}(\mathcal{D}, \theta) =\ & \ell_\text{SAM2} (\mathcal{D}, \theta^{\texttt{vgg}}, \theta^\texttt{(k)})
         + \ell_\text{ctrs}(\mathcal{D}, \theta_{pa}, \theta_{pv}),
    \end{aligned}
\end{equation}
where \( \theta^\texttt{(k)}=\{\theta_{pe}^\texttt{(k)}, \theta_{c}^\texttt{(k)}, \theta_{v}^\texttt{(k)}, \theta_{f}^\texttt{(k)}, \theta_{s}^\texttt{(k)}(\text{if }k\ge2) \mid k \in \{1, 2, 3\}\}\). During the optimisation, we only supervise the mask with the lowest segmentation loss in \(\ell_{\text{SAM2}}\).

\section{Experiment}
\begin{table*}[t!]
\centering
\caption{
\textbf{Comparison with SOTA on the Ref-AVS dataset}. Methods based on SAM~\cite{kirillov2023segment} are shown in \smash{\colorbox{lightgrape}{mauve}}, those based on SAM2~\cite{ravi2024sam} in \smash{\colorbox{yellow!10}{yellow}}, while other entries use task-specific models. The $\dagger$ indicates our reimplementation and * denotes methods utilising SAM’s zero-shot capability. The best results are marked in \textcolor{darkred}{red}, and the second best are underlined.}
\vspace{-7.5pt}
\label{tab:ref_avs}
\renewcommand{\arraystretch}{1.1}
\resizebox{.9\linewidth}{!}{
\begin{tabular}{?r?c?ccc?ccc?ccc?c?}
\specialrule{1.5pt}{0pt}{0pt}
\multicolumn{1}{?c?}{\multirow{3}{*}{Method}} & \multirow{3}{*}{Backbone} & \multicolumn{10}{c?}{Ref-AVS~\cite{wang2025ref}} \\
\cline{3-12}
                        &                           & \multicolumn{3}{c?}{Seen} & \multicolumn{3}{c?}{Unseen} & \multicolumn{3}{c?}{Mix}                             & \multicolumn{1}{c?}{Null}                                             \\
\cdashline{3-12}
                        &                                & $\mathcal{M_J \uparrow}$      & $\mathcal{M_F \uparrow}$ & $\mathcal{J\&F \uparrow}$          & $\mathcal{M_J\uparrow}$       & $\mathcal{M_F\uparrow}$ & $\mathcal{J\&F\uparrow}$     & $\mathcal{M_J\uparrow}$ & $\mathcal{M_F\uparrow}$ & $\mathcal{J\&F\uparrow}$ & $\mathcal{S\downarrow}$ \\
\specialrule{1.5pt}{0pt}{0pt}
TPAVI~\cite{zhou2022audio}~{\scriptsize \textcolor{gray}{[ECCV 2022]}}                & PVT-v2                             & 23.20  & 51.1       &   37.2          &32.36       &  54.7    &    43.5            & 27.78       &  52.9    &  40.3  &   0.208           \\

AVSegFormer~\cite{gao2023avsegformer}~{\scriptsize \textcolor{gray}{[AAAI 2024]}}             & PVT-v2                            & 33.47         &  47.0      &     40.2        & 36.05          & 50.1  &  43.1       & 34.76 & 48.6 & 41.7  & 0.171     \\ 
EEMC~\cite{wang2025ref}~{\scriptsize \textcolor{gray}{[ECCV 2024]}}             & Swin-b                           & 34.20               &  51.3        &   42.8        & 49.54          & 64.8  &  57.2       & 41.87 & 58.1  & 50.0 & \textcolor{darkred}{0.007}     \\
\specialrule{1.5pt}{0pt}{0pt}
\rowcolor{lightgrape!40}
GAVS~\cite{wang2024prompting}~{\scriptsize \textcolor{gray}{[AAAI 2024]}} & ViT-h & 28.9 &49.8 &39.35 &29.8 &49.7 & 39.8 &29.4 &49.8 & 39.6 &0.190 \\
\rowcolor{lightgrape!40}
SAMA-AVS~\cite{wang2024prompting}~{\scriptsize \textcolor{gray}{[WACV 2024]}}  & ViT-h & 39.2 &56.2 &47.7 &47.5 &56.6 & 52.1 & 43.4 & 56.4 & 49.9 & 0.130 \\
\rowcolor{lightgrape!40}
TSAM~\cite{radman2025tsam}~{\scriptsize \textcolor{gray}{[CVPR 2025]}} &  ViT-h & 43.4 & 56.8 & 50.1 & 54.6 & 66.4 & 60.5 & 49.0 & 61.6 & 55.3 & 0.017 \\
\rowcolor{lightgrape!60}
Ours~{\scriptsize \textcolor{gray}{SAM (w/ AuralFuser)}} &  ViT-h & 48.26 & 60.28 & 54.27 & 57.91 & 68.95 & 63.43 &  53.09 & 59.10 & 58.85 & 0.053 \\
\rowcolor{yellow!5}
GroundedSAM2$^*$~\cite{ren2024grounded}~{\scriptsize \textcolor{gray}{[arxiv 2024]}} & Hiera-b+ & 28.5 & 39.9 & 34.2 & 59.8 & 68.1 & 63.9 & 44.2 & 54.0 & 49.1 & 0.277 \\
\rowcolor{yellow!5}
GAVS$^\dagger$~\cite{wang2024prompting}~{\scriptsize \textcolor{gray}{[AAAI 2024]}}                    & Hiera-b+                            &    48.0          &  54.6              &    51.3    &       59.2        &    65.8        &       62.5        &   53.6           &        60.2  & 56.9  &    0.076                \\
\rowcolor{yellow!5}
SAMA-AVS$^\dagger$~\cite{wang2024prompting}~{\scriptsize \textcolor{gray}{[WACV 2024]}}                    & Hiera-b+                            &     49.5         & 56.7             &   53.1       &           60.6    &     66.4        &     63.5        &    55.1           &    61.5    &  58.3   &     0.103               \\
\rowcolor{yellow!5}
SAM2-LOVE~\cite{wang2025sam2}~{\scriptsize \textcolor{gray}{[CVPR 2025]}} & Hiera-l & 43.5 & 51.9 & 47.7 & \ul{66.5} & \ul{72.3} & \ul{69.4} & 55.0 & 62.1 & 58.5 & 0.230 \\
\rowcolor{yellow!10}
                    & Hiera-b+                         &       \ul{53.16}        &     \ul{58.83}       &     \ul{56.00}       &  63.45             &  70.44    &   66.95             &        \ul{58.31}       &   \ul{64.64}       &  \ul{61.48}  &    0.129                    \\
\rowcolor{yellow!10}
\multirow{2}{*}[3.3ex]{Ours~{\scriptsize \textcolor{gray}{AuralSAM2}}}                    & Hiera-l                   &            \textcolor{darkred}{56.16}           &     \textcolor{darkred}{61.19}           &   \textcolor{darkred}{58.68}   &   \textcolor{darkred}{68.69}            &    \textcolor{darkred}{74.36}             &  \textcolor{darkred}{71.53}   &          \textcolor{darkred}{62.43}     &   \textcolor{darkred}{ 67.78}     &   \textcolor{darkred}{65.11}  &   \ul{0.065}  \\
\specialrule{1.5pt}{0pt}{0pt}
\end{tabular}}
\vspace{-5pt}
\end{table*}

\begin{table*}[t!]
\centering
\caption{\textbf{Comparison with SOTA on the AVSBench dataset.} Methods employing SAM~\cite{kirillov2023segment} are in \smash{\colorbox{lightgrape}{mauve}}, SAM2~\cite{ravi2024sam} in \smash{\colorbox{yellow!10}{yellow}}, and the rest are task-specific models. The $\dagger$ denotes our reimplementation, $\#$ represents grounding semantic information to the class-agnostic mask via~\cite{ma2024steppingstones}, and $*$ denotes methods utilising SAM's zero-shot capability. The best results are in \textcolor{darkred}{red} and the second best are underlined.}
\vspace{-7.5pt}
\label{tab:avs}
\renewcommand{\arraystretch}{1.1}
\resizebox{.9\linewidth}{!}{
\begin{tabular}{?r?c?cc?cc?cc|cc?}
\specialrule{1.5pt}{0pt}{0pt}

\multicolumn{1}{?c?}{\multirow{3}{*}{Method}} & \multirow{3}{*}{ \hspace{-8pt}\shortstack[c]{ Backbones \\ \\ \text{\small (audio \& visual)}}} & \multicolumn{8}{c?}{AVSBench~\cite{zhou2022audio, zhou2023audio}} \\
\cline{3-10}
                        &                           & \multicolumn{2}{c?}{V1{\small (single)} } & \multicolumn{2}{c?}{V1{\small (multiple)}} & \multicolumn{2}{c|}{V2 (binary)} & \multicolumn{2}{c?}{V2 (semantic)}                                             \\
\cdashline{3-10}
                        &                                & $\mathcal{M_J\uparrow}$      & $\mathcal{M_F\uparrow}$           & $\mathcal{M_J\uparrow}$       & $\mathcal{M_F\uparrow}$      & $\mathcal{M_J\uparrow}$ & $\mathcal{M_F\uparrow}$ & $\mathcal{M_J\uparrow}$ & $\mathcal{M_F\uparrow}$ \\
\specialrule{1.5pt}{0pt}{0pt}
AVSegFormer~\cite{gao2023avsegformer}~{\scriptsize \textcolor{gray}{[AAAI 2024]}}             & \makebox[2.5cm][l]{\text{\small VGGish} \hspace{2pt} PVT-v2}                            & 82.1         & 89.9                   & 58.4          & 69.3           & \multicolumn{2}{c|}{-}    & 36.7 & 42.0    \\
AVS-BiGen~\cite{hao2023improving}~{\scriptsize \textcolor{gray}{[AAAI 2024]}}               & \makebox[2.5cm][l]{\text{\small VGGish} \hspace{2pt} PVT-v2}                             & 81.7         & 90.4                    & 55.1          & 66.8                  & 64.3          & 75.9            & \multicolumn{2}{c?}{-}            \\
Step.-Stones~\cite{ma2024steppingstones}~{\scriptsize \textcolor{gray}{[ECCV 2024]}}                      & \makebox[2.5cm][l]{\text{\small VGGish} \hspace{2pt} Swin-b}                             & 83.2       &91.3                      &  67.3      & 77.6             & \multicolumn{2}{c|}{-}        & 48.5$^\#$ & 53.2$^\#$        \\
\specialrule{1.2pt}{0pt}{0pt}
\rowcolor{lightgrape!40}
SAM4AVS$^*$~\cite{yu2023can}~{\scriptsize \textcolor{gray}{[BMVC 2023]}}   & \makebox[2.5cm][l]{\text{\small VGGish} \hspace{2pt} PVT-v2}                              & 51.2         &  61.5                  &   41.8        &  47.8                   &     \multicolumn{2}{c|}{-}          &          \multicolumn{2}{c?}{-}           \\  
\rowcolor{lightgrape!40}
COMBO$^*$~\cite{yang2023cooperation}~{\scriptsize \textcolor{gray}{[CVPR 2024]}}                & \makebox[2.5cm][l]{\text{\small VGGish} \hspace{2pt} PVT-v2}                                 & 84.7         & 91.9                   & 59.2          & 71.2                   &      \multicolumn{2}{c|}{-}    &    42.1           & 46.1         \\
\rowcolor{lightgrape!40}
GAVS~\cite{wang2024prompting}~{\scriptsize \textcolor{gray}{[AAAI 2024]}}                     & \makebox[2.5cm][l]{\text{\small VGGish} \hspace{2pt} ViT-b}                             & 80.1         & 90.2                   & 63.7          & 77.4                     & 67.7          & 78.8             &  \multicolumn{2}{c?}{-}             \\
\rowcolor{lightgrape!40}
SAMA-AVS~\cite{liu2024annotation}~{\scriptsize \textcolor{gray}{[WACV 2024]}}   &  \makebox[2.5cm][l]{\text{\small VGGish} \hspace{2pt} ViT-h}                              & 81.5         & 88.6                   & 63.1          &  69.1                    &        \multicolumn{2}{c|}{-}           &         \multicolumn{2}{c?}{-}            \\   
\rowcolor{lightgrape!60}
Ours~{\scriptsize \textcolor{gray}{SAM (w/ AuralFuser)}}   &  \makebox[2.5cm][l]{\text{\small VGGish} \hspace{2pt} ViT-h}                              & 84.78         & 91.92                   & 65.22          &  79.13                    &        70.24           & 81.63 &      49.52$^\#$ & 54.88$^\#$            \\   
\rowcolor{yellow!5}
AL-Ref$^*$~\cite{huang2024unleashing}~{\scriptsize \textcolor{gray}{[AAAI 2025]}}                    &  \makebox[2.5cm][l]{\hspace{2pt} Beats \hspace{5pt} Hiera-l}                         &       70.5        &     81.1                   &  48.6             &  53.5                        &   59.2           &   66.2     &        36.0       &   39.8            \\
\rowcolor{yellow!5}
GAVS$^\dagger$~\cite{wang2024prompting}~{\scriptsize \textcolor{gray}{[AAAI 2024]}}                    & \makebox[2.5cm][l]{\text{\small VGGish} \hspace{2pt} Hiera-b+}                            &   83.64           &   \ul{92.47}                     &           68.13    & 79.07                         &      73.58         &    84.04         &           \multicolumn{2}{c?}{-}           \\
\rowcolor{yellow!5}
SAMA-AVS$^\dagger$~\cite{liu2024annotation}~{\scriptsize \textcolor{gray}{[WACV 2024]}}                    & \makebox[2.5cm][l]{\text{\small VGGish} \hspace{2pt} Hiera-b+}                             &     82.11        &  90.58                     &    67.70    &  78.93                        &   74.28       & 84.35          &           \multicolumn{2}{c?}{-}              \\
\rowcolor{yellow!10}
                  & \makebox[2.5cm][l]{\text{\small VGGish} \hspace{2pt} Hiera-b+}                         &       \ul{85.01}        &     92.16                   &  \ul{72.04}             &  \ul{81.46}                     &        \ul{76.78}       &   \ul{85.38}          &       \ul{50.23$^\#$}       &  \ul{55.16$^\#$}            \\
\rowcolor{yellow!10}
\multirow{2}{*}[3.3ex]{Ours~{\scriptsize \textcolor{gray}{AuralSAM2}}}  & \makebox[2.5cm][l]{\text{\small VGGish} \hspace{2pt} Hiera-l}                   &            \textcolor{darkred}{86.62}           &     \textcolor{darkred}{93.34}                  &   \textcolor{darkred}{75.58}            &    \textcolor{darkred}{84.12}                  &          \textcolor{darkred}{79.09}     &   \textcolor{darkred}{86.84}          &\textcolor{darkred}{50.57$^\#$}           &  \textcolor{darkred}{56.03$^\#$} \\
\specialrule{1.5pt}{0pt}{0pt}
\end{tabular}
}
\vspace{-10pt}
\end{table*}

\noindent \textbf{Experimental setup.} With language-aided AVS, we evaluate our method on \textbf{\textit{Ref-AVS}}~\cite{wang2025ref} benchmark, which includes 4,002 video clips and 20,261 expressions. Each expression corresponds to a unique object, with 14,117 training and 4,770 test cases. The test set is divided into 2,288 seen-object cases for performance evaluation, 1,454 unseen-object cases for generalisation assessment, and 1,028 null cases where the referenced object is absent or not visible.
We also evaluate our method on the \textbf{\textit{AVSBench}}~\cite{zhou2022audio} dataset without language modality, which comprises two subsets: \textit{V1s} and \textit{V1m}, representing single and multiple sounding sources, respectively. The \textit{V1s} subset consists of 3,452 training clips, 740 validation clips, and 740 test clips, while the \textit{V1m} subset includes 296 training cases, 64 validation cases, and 64 test cases, both evaluated in a binary class-agnostic setting. The extended \textit{V2}~\cite{zhou2023audio} subset builds upon \textit{V1s} and \textit{V1m}, introducing 12,356 video clips across 70 semantic categories.
\begin{table*}[t!]
\caption{\textbf{Ablation Studies} on AVSBench and Ref-AVS using Hiera~\cite{ryali2023hiera} large backbones. \colorbox{gray!10}{The first row} presents results based solely on the visual modality \visualiconincap, while the following rows show outcomes from cross-modal fusion with audio \audioiconincap~ or optional language \texticonincap~ modalities. The subsequent two rows illustrate the effect of employing a multi-scale feature pyramid \pyramidiconincap~ arranged from bottom to up, with the \colorbox{yellow!10}{bottom row} further incorporating audio-guided contrastive learning.}
\vspace{-7.5pt}
\label{tab:ablation_avs}
\renewcommand{\arraystretch}{1.1}
\centering
\resizebox{\linewidth}{!}{
\begin{tabular}{?c?c?ccc?ccc?ccc?ccc?}
\specialrule{1.5pt}{0pt}{0pt}
\multirow{3}{*}{Ablations} & \multirow{3}{*}{Pyramid}                     & \multicolumn{6}{c?}{AVSBench~\cite{zhou2022audio, zhou2023audio}} & \multicolumn{6}{c?}{Ref-AVS~\cite{wang2025ref}}                    \\
\cline{3-14}
                         &                          &  \multicolumn{3}{c?}{V1 (single)} & \multicolumn{3}{c?}{V1 (multiple)}  & \multicolumn{3}{c?}{Seen} & \multicolumn{3}{c?}{Unseen}\\
\cdashline{3-14}
                           &                          &  $\mathcal{M_J}\uparrow$     &  $\mathcal{M_F}\uparrow$    &  $\mathcal{J\&F}\uparrow$    &  $\mathcal{M_J}\uparrow$     &  $\mathcal{M_F}\uparrow$    &  $\mathcal{J\&F}\uparrow$      &  $\mathcal{M_J}\uparrow$     &  $\mathcal{M_F}\uparrow$    &  $\mathcal{J\&F}\uparrow$ &  $\mathcal{M_J}\uparrow$     &  $\mathcal{M_F}\uparrow$    &  $\mathcal{J\&F}\uparrow$   \\
\specialrule{1.5pt}{0pt}{0pt}
\rowcolor{gray!10}
            \visualicon                          & -                      & 83.41       &  91.36     &  87.39   & 61.50     & 73.09       & 67.30 &  43.77   &  48.01         &  45.89      & 64.33 & 69.98    & 67.16 \\
 \audioicon      \hspace{3pt}  (\texticon)   \visualicon                          & -                      &   84.55     &  92.08     &   88.32     & 71.52     & 79.57       &    75.55             &  53.36      &  57.49   &    55.43    & 66.94 & 72.18    & 69.56 \\
 \audioicon      \hspace{3pt}  (\texticon)  \visualicon                          & \pyramidicon=2                      &    85.96    &  92.97     &  89.47      &   73.42    &   81.94   &        77.68       &    54.67    & 58.91    &    56.79    & 67.81 &72.86     & 70.34 \\
\audioicon   \hspace{3pt}  (\texticon)         \visualicon                       & \pyramidicon=3                      &     86.33   &   93.27    &   89.80     &    74.43    &   82.76    &    78.60     &  55.32       & 60.69    &  58.00     &67.74   & 73.92    & 70.83\\
\rowcolor{yellow!10}
\audioicon \hspace{3pt} (\texticon)  \visualicon \hspace{3pt} AudioCon             & \pyramidicon=3                      &     86.62   &    93.34    & 89.98  &  75.58      &  84.12     &  79.85    &  56.16  &  61.19    &  58.68   &   68.69      & 74.36       &   71.53          \\    
\specialrule{1.5pt}{0pt}{0pt}
\end{tabular}}
\vspace{-10pt}
\end{table*}

\vspace{-9pt}
\noindent \textbf{Metrics.} We use the average Jaccard index (\(\mathcal{M_J}\)) and F-Score (\(\mathcal{M_F}\)) for evaluating segmentation performance in AVSBench~\cite{zhou2022audio}, along with an additional Square Root of the Ratio measurement (\(\mathcal{S}\)) in Ref-AVS~\cite{wang2025ref}. \\
\textbf{Implementation Details}. Our experiments are built upon the SAM2 framework~\cite{ravi2024sam} using both the Hiera\_base+ and Hiera\_large backbones. Following previous SAM-based methods~\cite{wang2024prompting}, we use an input image resolution of 1024x1024 and a batch size of one across all datasets. Given the limited exploration of SAM2 within AVS, we have re-implemented previous SOTA methods~\cite{wang2024prompting, liu2024annotation} based on their code. During training, the learning rate is set to 1e$^{-4}$, with a poly learning rate decay following \((1 - \frac{\text{iter}}{\text{max iter}})^{0.9}\). Consistent with SAM2~\cite{ravi2024sam}, we set \(20:1:1:1\) for the linear combination for \(\ell_{\text{focal}},\ell_{\text{dice}}, \ell_{\text{IoU}}\) and \(\ell_{\text{occ}}\) in Eq.~\eqref{eq:sam2_loss}. For contrastive learning, a three-layer projector is used for both audio and visual features, with an output dimension of 64. The temperature value is set to \(\tau=0.10\) in Eq.~\eqref{eq:ctrs_loss} and remains constant throughout all experiments. 
Please refer to Supp. Section~\textcolor{red}{1} for more implementation details and to Supp. Section~\textcolor{red}{3} for results with other backbones. 
\subsection{Comparing with SOTA Methods}
\label{sec:exp_sota_tables}
\noindent \textbf{Results on Ref-AVS Dataset.} As shown in Tab.~\ref{tab:ref_avs}, we evaluate our method on an audio-language-visual task. With the Hiera\_base+ backbone, our approach outperforms GAVS~\cite{wang2024prompting} by 5.2\% in Jaccard for seen scenarios, demonstrating an enhanced ability to integrate complex multi-modalities. 
Further, upgrading to the Hiera\_l backbone yields an average Jaccard improvement of 4.12\% compared to Hiera\_base+, as detailed in the 'Mix' rows.  Additionally, our method (AuralFuser) can be directly integrated into SAM~\cite{ke2024segment}, improving over TSAM~\cite{radman2025tsam} by 4.17\% on the Seen average, highlighting its strong generalisation.
\\
\noindent \textbf{Results on AVS Datasets.} In Tab.~\ref{tab:avs}, we evaluate our approach on the AVSBench dataset under the audio-visual setting.
With the Hiera\_base+ backbone, our method surpasses adapter-based counterparts, achieving a 4.34\% Jaccard gain over SAMA-AVS~\cite{liu2024annotation} on the V1m subset, highlighting the effectiveness of our cross-modal fusion design. 
Moreover, our method outperforms the zero-shot baseline~\cite{huang2024unleashing} by 22.8\%, demonstrating that our feature-level prompts provide stronger guidance to SAM2 than external foundation models. Our method also improves the SAM~\cite{ke2024segment} architecture; for example, it boosts performance on the V1m subset by 2.12\% over SAMA-AVS~\cite{liu2024annotation} and 1.52\% over GAVS~\cite{wang2024prompting}, effectively mitigating the audio prompt dilution issue across different SAM families.
\begin{table}[]
\caption{\textbf{Ablation Studies on CL} in AVSBench~\cite{zhou2022audio} dataset based on Hiera\_l backbone. 
Best results are highlighted in \textcolor{darkred}{red}. }
\vspace{-10pt}
\label{tab:contrastive}
\centering
\renewcommand{\arraystretch}{1.1}
\resizebox{\linewidth}{!}{
\begin{tabular}{?c?cccc?}
\specialrule{1.5pt}{0pt}{0pt}
\renewcommand{\arraystretch}{1.1}
\multirow{2}{*}{Method} & \multicolumn{4}{c?}{AVSBench (V1m)}  \\
\cline{2-5}
                        & \(\mathcal{M_J}\uparrow\)      & \(\mathcal{M_F}\uparrow\)     & \(\mathcal{J\&F}\uparrow\)  &  \(\mathbf{\Delta}\) \\
\hline
w/o CL       &      74.43        &    82.76          &   78.60     & -   \\
Ours w/  SupCon         &    74.86          &   83.29          & 79.08 & {\textcolor{gray}{0.48 \(\uparrow\)}}         \\
\rowcolor{yellow!10}
Ours w/ AudioCon        &     \textcolor{darkred}{75.58}         &     \textcolor{darkred}{84.12}        &  \textcolor{darkred}{79.85} & {\textcolor{gray}{1.25 \(\uparrow\)}}      \\
\specialrule{1.5pt}{0pt}{0pt}
\end{tabular}}
\vspace{-10pt}
\end{table}

\begin{table}[t!]
\caption{\textbf{Ablation study of prompts} on AVSBench~\cite{zhou2022audio} using the Hiera\_l backbone. Best results are shown in \textcolor{darkred}{red}.}
\vspace{-7.5pt}
\label{tab:prompt_ablation}
\renewcommand{\arraystretch}{1.1}
\centering
\resizebox{\linewidth}{!}{
\begin{tabular}{!{\vrule width 1.pt}c!{\vrule width 1.15pt}cccr!{\vrule width 1.15pt}}
\specialrule{1.5pt}{0pt}{0pt}
\multirow{2}{*}{Method} & \multicolumn{4}{c!{\vrule width 1.15pt}}{AVSBench (V1m)} \\
\cline{2-5}
                        & \(\mathcal{M_J}\uparrow\)      & \(\mathcal{M_F}\uparrow\)     & \(\mathcal{J\&F}\uparrow\)  & \multicolumn{1}{c!{\vrule width 1.15pt}}{ \(\mathbf{\Delta}\)} \\
\hline
\rowcolor{yellow!10}
Ours                & \textcolor{darkred}{75.58}         &     \textcolor{darkred}{84.12}        &  \textcolor{darkred}{79.85} & \multicolumn{1}{c!{\vrule width 1.15pt}}{-}\\
w/o Sparse Prompts              &       67.06   &  76.51        &  71.79  &   {\textcolor{gray}{8.06 \(\downarrow\)}}         \\
w/o Dense Prompts               &      63.32    &  73.15        & 68.24  &   {\textcolor{gray}{11.61 \(\downarrow\)}}          \\
\specialrule{1.5pt}{0pt}{0pt}
\end{tabular}}
\vspace{-12pt}
\end{table}

\begin{figure}
    \centering
    \includegraphics[width=.9\linewidth]{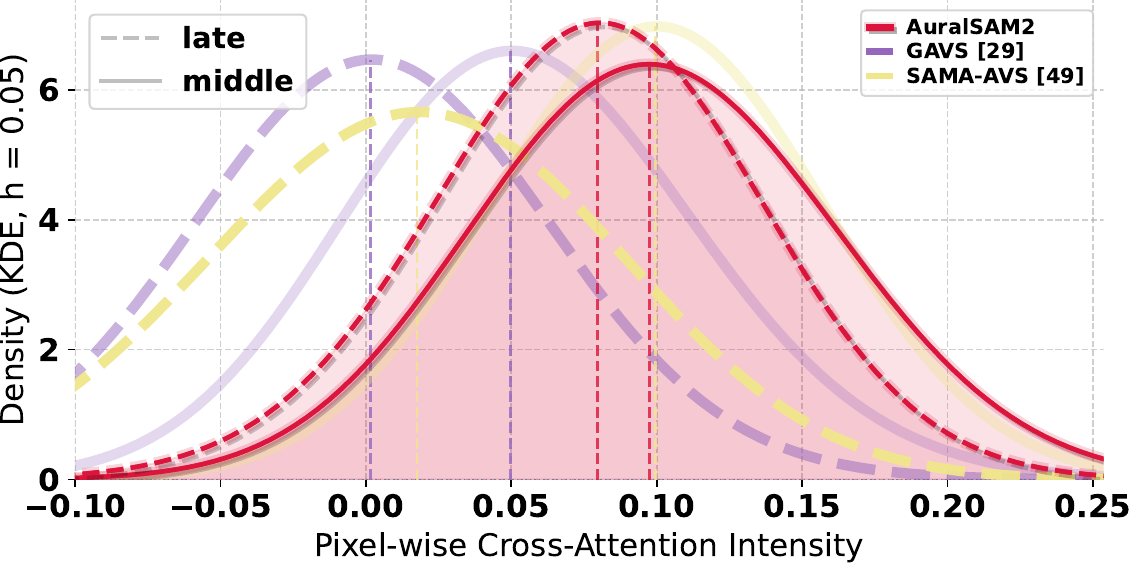}
    \vspace{-10pt}
\caption{\textbf{PDF of cross-attention intensity} between audio cues and pixels on AVSBench (V1m)~\cite{zhou2022audio}. Values indicate pixel-wise cross-attention between the audio prompt and visual features.}

\vspace{-10pt}
    \label{fig:pdf}
\end{figure}
\subsection{Ablation Studies}
\begin{figure*}[t!]
    \centering
    \includegraphics[width=.90\linewidth]{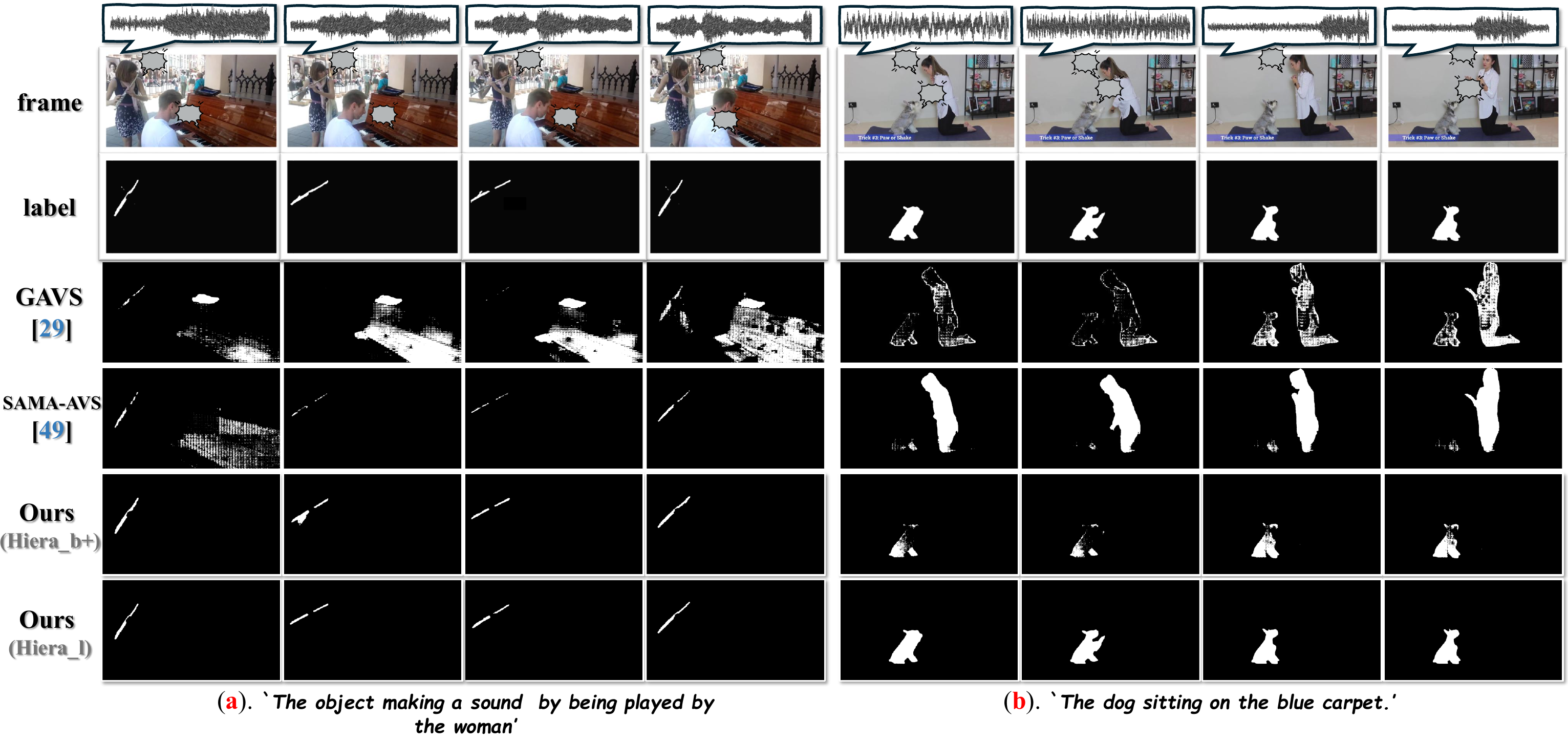} 
    \vspace{-10pt}
    \caption{Qualitative visualisations on the Ref-AVS~\cite{wang2025ref} dataset. The first row shows the input frame, followed by the ground truth labels in the second row. The third and fourth rows present adaptor-based methods~\cite{wang2024prompting, liu2024annotation} using the SAM2 architecture with the Hiera\_b+ backbone, while our method is displayed in the last two rows. Please refer to Supp. Section \textcolor{red}{4} for additional qualitative results.}
    \label{fig:visualisation}
    \vspace{-15pt}
\end{figure*}

\label{sec:exp_ablation_tables}
\noindent We summarize component-wise performance gains in Tab.~\ref{tab:ablation_avs}. The first row shows the baseline using only the visual modality. Incorporating audio and language in Ref-AVS\cite{wang2025ref} improves $\mathcal{J\&F}$ by 8.25\% on AVSBench (V1m) and 9.54\% on the Seen subset of Ref-AVS. Adding the feature pyramid further boosts performance by 3.55\% and 2.57\% on the respective datasets, demonstrating its effectiveness in capturing richer semantics for cross-modal fusion. Finally, introducing AudioCon improves results by another 1.25\% and 0.84\%, enhancing the alignment between vision and other modalities. \\
\noindent\textbf{Probability Density of Cross-Attention Intensity} across the network on AVSBench (V1m)~\cite{zhou2022audio}, shown in Fig.~\ref{fig:pdf}.
In later attention layers, our method exhibits a higher density centered around 0.075, while SAMA-AVS~\cite{liu2024annotation} peaks near 0.01, indicating that our approach effectively mitigates audio prompt dilution and enables stronger prompting. Notably, the shift in density modes between mid and late stages is smaller for our method, suggesting more consistent audio–visual alignment across the network propagation. \\
\noindent \textbf{Ablation Studies on Feature-Level Prompts}. As shown in Tab.\ref{tab:prompt_ablation}, we evaluate the importance of feature-level prompts by omitting them one at a time in Eq.~\eqref{eq:prompting} on AVSBench (V1m) with the Hiera\_l backbone. The results indicate that both are essential to our module; for example, removing sparse prompts reduces the \(\mathcal{J\&F}\) score by 8.06\%, while removing dense prompts decreases it by 11.61\%. \\
%
\begin{table}[]
  \caption{\textbf{Prompt Engineering with Audio} in the AVSBench (V1m)~\cite{zhou2022audio} dataset with Hiedra\_base+ backbone. We use points and boxes generated from ground truth to simulate real-world prompting practices. The FPS represents the number of frames processed per second, and the best results highlighted in \textcolor{darkred}{red}.}
    \label{tab:prompt_engineering}
\vspace{-7.5pt}
\centering
\renewcommand{\arraystretch}{1.2}
\resizebox{.9\linewidth}{!}{
\begin{tabular}{?r|c?cc|c?}
\specialrule{1.2pt}{0pt}{0pt}
\multicolumn{1}{?c?}{Methods}              & Prompts            & \(\mathcal{M_J}\) & \(\mathcal{M_F}\) & FPS  \\
\hline
\multicolumn{1}{?c?}{\multirow{4}{*}{SAM2~\cite{ravi2024sam}}} & points             & 64.67   & 72.15   & 17.8           \\
                      & box                & 68.85   & 76.52   & 17.4          \\
                      
                      & \cellcolor{gray!10} mask               & \cellcolor{gray!10} 75.73   & \cellcolor{gray!10} 81.54   & \cellcolor{gray!10} 16.9          \\
                      & points box         & 72.64   & 79.56   & 17.2          \\
\cdashline{1-5}
\rowcolor{yellow!10}
GAVS~\cite{wang2024prompting} {\scriptsize (w/ SAM2)}             & audio points box & 71.70   & 81.94   & 8.7          \\
\rowcolor{yellow!10}
SAMA-AVS~\cite{liu2024annotation} {\scriptsize (w/ SAM2)}         & audio points box & 69.74   & 80.97   &   9.9        \\
\rowcolor{yellow!10}
Ours {\scriptsize (w/ SAM2)}     & audio points box & \textcolor{darkred}{74.26}   & \textcolor{darkred}{83.58}   & \textcolor{darkred}{14.1}   \\
\specialrule{1.2pt}{0pt}{0pt}     
\end{tabular}
}
\vspace{-15pt}
\end{table}


\vspace{-5pt}
\label{sec:exp_prompting_tables}
\noindent \textbf{Ablation Studies on CL.} In Tab.~\ref{tab:contrastive}, we present ablation studies on contrastive learning using the AVSBench (V1m)~\cite{zhou2022audio} dataset. The first row reports results without CL, the second row applies SupCon~\cite{khosla2020supervised}, originally designed for vision-only tasks, and the last row showcases our proposed AudioCon. Our method achieves an additional 0.77 \(\mathcal{J\&F}\) improvement over SupCon in AVS, demonstrating superior audio-visual alignment. \\
\textbf{Promptable segmentation with SAM2.} In Tab.~\ref{tab:prompt_engineering}, we simulate prompt engineering in a human-in-the-loop setting on AVSBench (V1m)~\cite{zhou2022audio}. The visual prompts are derived from the ground truth, consisting of four uniformly generated points per frame along with the corresponding bounding box, applied to the first frame following the SAM2 inference pipeline. Since preserving \colorbox{gray!10}{pixel-level labelled masks} in practice is challenging, we use only points and boxes in this experiment. 
As a result, compared to other adapter-based methods~\cite{wang2024prompting, liu2024annotation}, our approach achieves the best performance in both measurements. For example, it increases Jaccard by 2.56\% over GAVS~\cite{wang2024prompting} while maintains high efficiency with 14.1 frame-per-second (FPS) throughput.
\vspace{-2.5pt}
\subsection{Visualisation}
\vspace{-2.5pt}
We present qualitative results on Ref-AVS~\cite{wang2025ref} in Fig.~\ref{fig:visualisation}, where our method shows superior visual performance. In case (\textcolor{red}{a}), given the expression \texttt{\small `the object making a sound by being played by the woman'}, prior methods~\cite{wang2024prompting, liu2024annotation} either misidentify the \texttt{\small piano} or fail to accurately segment the thick \texttt{\small flute} accurately. In contrast, our approach precisely captures the flute, achieving higher accuracy with the Hiera\_l backbone. \\
\label{sec:exp_visualisation}
\vspace{-5pt}

\vspace{-12pt}
\section{Conclusion}
We introduce AuralSAM2, a novel framework that enables SAM2 to process audio without relying on adapters or external foundation models. To address the inefficiency of repeated inference, we propose AuralFuser, a module that integrates multimodal features and directly generates sparse and dense feature-level prompts. These prompts guide the decoder without modifying image features, preserving SAM2’s efficiency and generalizability in promptable segmentation. To mitigate audio prompt dilution, AuralFuser performs cross-modal fusion within a multi-scale feature pyramid, enhancing both contextual understanding and fine-grained alignment. Finally, to alleviate visual dominance in the latent space caused by the imbalance between visual and audio embeddings, we introduce AudioCon, which promotes alignment around audio signals as semantic anchors.

{
    \small
    \bibliographystyle{ieeenat_fullname}
    \bibliography{main}
}


\end{document}